\documentclass[10pt,twocolumn,letterpaper]{article}
\usepackage{iccv}
\usepackage{times}
\usepackage{epsfig}
\usepackage{graphicx}
\usepackage{amsmath}
\usepackage{amssymb}
\graphicspath{{./fig/}}
\usepackage{xcolor}
\usepackage{caption}
\usepackage{subcaption}
\usepackage[ruled]{algorithm2e}
\usepackage{color}
\usepackage{enumerate}

\usepackage{makecell}

\usepackage[pagebackref=true,breaklinks=true,letterpaper=true,colorlinks,bookmarks=false]{hyperref}

\iccvfinalcopy 

\ificcvfinal\fi
\begin{document}
	
	\title{Clustered Object Detection in Aerial Images}
	
	
	\author{\;\; Fan Yang$^{1}$  \;\;   Heng Fan$^{1}$\;\; Peng Chu$^{1}$ \;\; Erik Blasch$^{2}$ \;\; Haibin Ling$^{3,1}$\thanks{Corresponding author.} \\
		{\normalsize $^{1}$Department of Computer and Information Sciences, Temple University, Philadelphia, USA}\\
		{\normalsize $^{2}$Air Force Research Lab, USA} \\
		{\normalsize $^{3}$Department Computer Science, Stony Brook University, Stony Brook, NY, USA.}\\
		{\tt\small \{fyang,hengfan,pchu\}@temple.edu, erik.blasch@us.af.mil, hling@cs.stonybrook.edu}
	}
	
	\maketitle
	\thispagestyle{empty}

	\begin{abstract}
		Detecting objects in aerial images is challenging for at least two reasons: (1) target objects like pedestrians are very small in pixels, making them hardly distinguished from surrounding background; and (2) targets are in general sparsely and non-uniformly distributed, making the detection very inefficient. In this paper, we address both issues inspired by observing that these targets are often clustered. In particular, we propose a Clustered Detection (ClusDet) network that unifies object clustering and detection in an end-to-end framework. The key components in ClusDet include a cluster proposal sub-network (CPNet), a scale estimation sub-network (ScaleNet), and a dedicated detection network (DetecNet). Given an input image, CPNet produces object cluster regions and ScaleNet estimates object scales for these regions. Then, each scale-normalized cluster region is fed into DetecNet for object detection. ClusDet has several advantages over previous solutions: (1) it greatly reduces the number of chips for final object detection and hence achieves high running time efficiency, (2) the cluster-based scale estimation is more accurate than previously used single-object based ones, hence effectively improves the detection for small objects, and (3) the final DetecNet is dedicated for clustered regions and implicitly models the prior context information so as to boost detection accuracy. The proposed method is tested on three popular aerial image datasets including VisDrone, UAVDT and DOTA. In all experiments, ClusDet achieves promising performance in comparison with state-of-the-art detectors. Code will be available in \url{https://github.com/fyangneil}.
	\end{abstract}
	
	\section{Introduction}

	\begin{figure}[t]
		\centering
		\includegraphics[width=\linewidth]{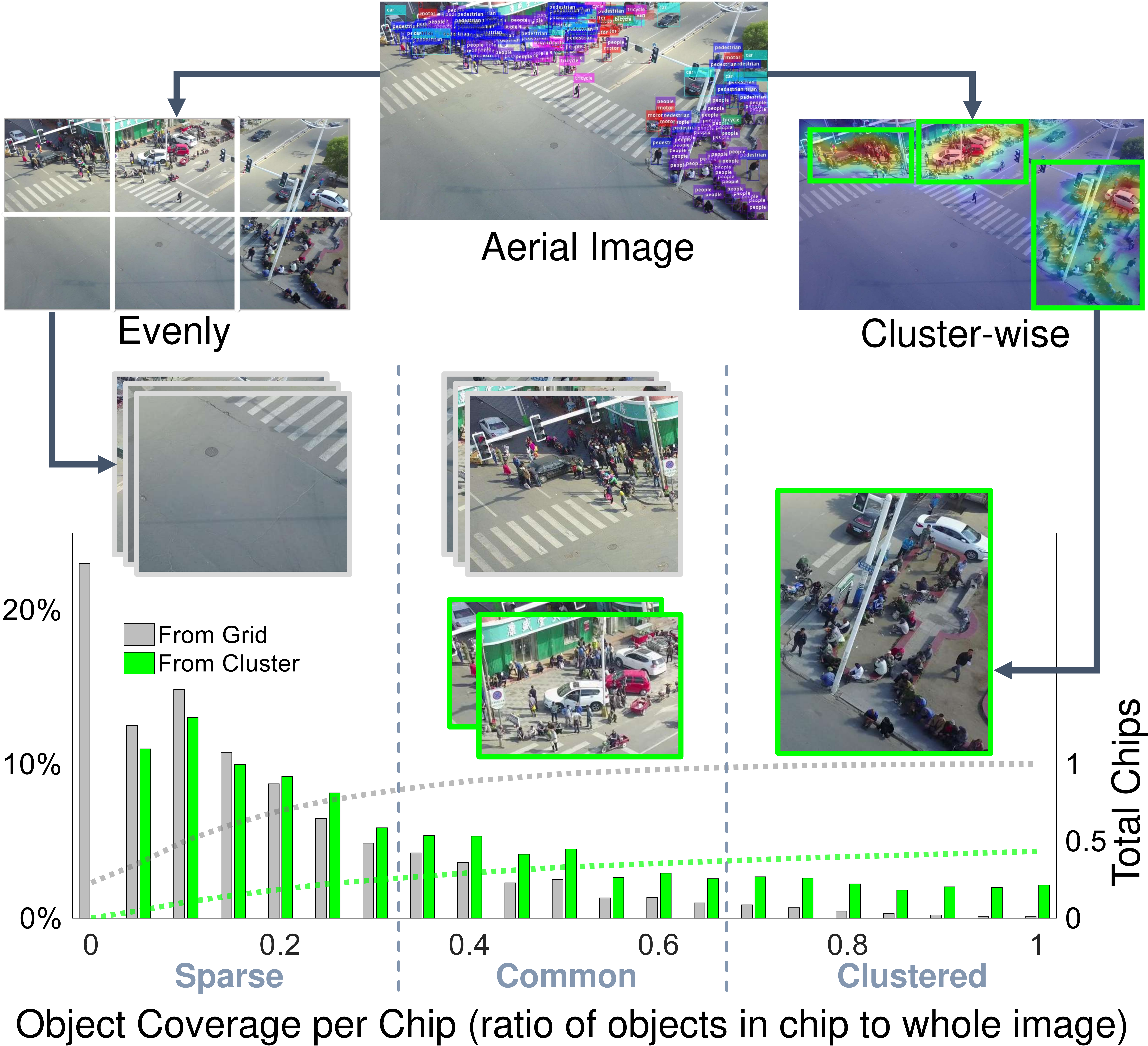}
		\caption{Comparison of grid-based uniform partition and the proposed cluster-based partition. For the narrative purpose, we intentionally classify a chip into three types: {\em sparse},  {\em common}, and {\em clustered}. We observe that, for grid-based uniform partition, more than 73\% chips are {\em sparse} (including 23\% chips with {\em zero} objects), around 25\% chips are {\em common}, and about 2\% chips are {\em clustered}. By contrast, for cluster-based partition, around 50\% chips are {\em sparse}, 35\% are {\em common}, and about 15\% belong to {\em clustered} chips, which is 7$\times$ more than that of grid-based partition.}
		\label{fig:grid_cluster}
	\end{figure}
	
	With the advance of deep neural networks, object detection (\eg, Faster R-CNN~\cite{ren2015faster}, YOLO~\cite{redmon2016you}, SSD~\cite{liu2016ssd}) has witnessed great progress for natural images (\eg, 600$\times$400 images in MS COCO~\cite{lin2014microsoft}) in recent years. Despite the promising results for general object detection, the performance of these detectors on the aerial images (\eg, 2,000$\times$1,500 pixels in VisDrone~\cite{zhu2018vision}) are far from satisfactory in both accuracy and efficiency, which are caused by two challenges: (1) targets typically have small scales relative to the images; and (2) targets are generally sparsely and non-uniformly distributed in the whole image.

	\begin{figure*}[t]
		\centering
		\includegraphics[width=\linewidth,height=.253\linewidth]{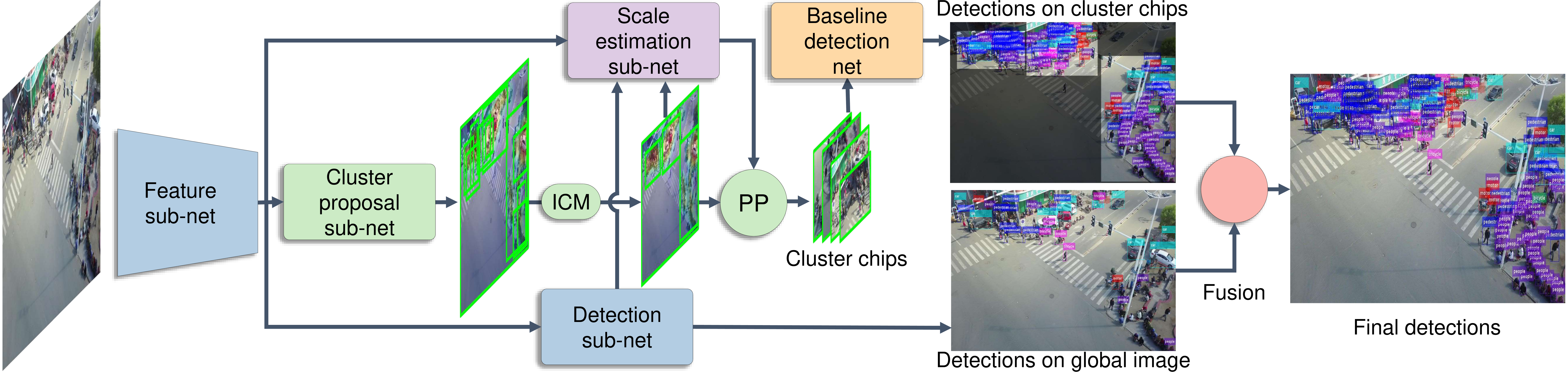}
		\caption{Clustered object Detection (ClusDet) network. The ClusDet network consists of three key components: (1) a cluster proposal subnet (CPNet); (2) a scale estimation subnet (ScaleNet); and (3) a dedicated detection network (DetecNet). CPNet serves to predict the cluster regions. ScaleNet is to estimate the object scale in the clusters. DetecNet performs detection on cluster chips. The final detections are generated by fusing detections from cluster chips and global image. The details of ICM (iterative cluster merging) and PP (partition and padding) are given in Section~\ref{method}.}
		\label{fig:cluster_aware_scheme}
	\end{figure*}
	
	Compared with objects in natural images, the scale challenge causes less effective feature representation of deep networks for objects in aerial images. Therefore, it is difficult for the modern detectors to effectively leverage appearance information to distinguish the objects from surrounding background or similar objects. In order to deal with the scale issue, a natural solution is to partition an aerial image into several uniform small chips, and then perform detection on each of them~\cite{gao2018dynamic,lu2016adaptive}. Although these approaches alleviate the resolution challenge to some extent, they are {\em inefficient} in performing detection due to the ignorance of the target sparsity. Consequently, a lot computation resources are inefficiently applied on regions with sparse or even no objects (see Fig.~\ref{fig:grid_cluster}). We observe from Fig.~\ref{fig:grid_cluster} that, in an aerial image objects are not only {\em sparse} and {\em non-uniform} but also tend to be highly {\em clustered} in certain regions. For example, pedestrians are usually concentrated in squares and vehicles on highways. Hence, an intuitive way to improve detection efficiency is to focus the detector on these clustered regions where there are a  large amount of objects.
	
	Inspired by this motivation, this paper proposes a novel clustered detection (ClusDet) network for addressing both challenges aforementioned by integrating object and cluster detection in a uniform framework.  As illustrated in Fig.~\ref{fig:cluster_aware_scheme}, ClusDet consists of three key components including a cluster proposal sub-network (CPNet), a scale estimation sub-network (ScaleNet) and a baseline detection network (DetecNet). According to the initial detection of an aerial image, CPNet generates a set of regions of object clusters. After obtaining the clustered regions, they are cropped out for subsequent fine detection. To such end, these regions have to be firstly resized to fit the detector, which may result in extremely large or small objects in the clustered regions and thus deteriorate the detection performance~\cite{singh2018analysis}. To handle this issue, we present the ScaleNet to estimate an appropriate scale for the objects in each cluster chip and then rescale the chip accordingly before feeding it to a detector, which is different from~\cite{gao2018dynamic,lu2016adaptive,lalonde2018clusternet} by directly resizing cropped chips. Afterwards, each clustered chip is fed to the dedicated detector, DetecNet, for fine detection. The final detection is achieved by fusing the detection results on both cluster chips and the global image. 
	
	Compared to previous approaches, the proposed ClusDet shows several advantages: (\romannumeral1) Owing to the CPNet, we only need to deal with the clustered regions with plenty of objects, which significantly reduces the computation cost and improves detection efficiency; (\romannumeral2) With the help of the ScaleNet, each clustered chip is refined for better subsequent fine detection, leading to improvement in accuracy; and (\romannumeral3) The DetecNet is specially designated for clustered region detection and implicitly models the prior context information to further boost detection accuracy. In extensive experiments on three aerial image datasets, ClusDet achieves the best performance using a single mode while with less computation cost. 
	
	In summary, the paper has the following contributions:
	\vspace{-.7em}
	\begin{enumerate}[1)]
		\setlength{\itemsep}{1pt}
		\setlength{\parsep}{0pt}
		\setlength{\parskip}{0pt}
		
		\item Proposes a novel ClusDet network to simultaneously address the scale and sparsity challenges for object detection in aerial images.
		
		\item Presents an effective ScaleNet to alleviate nonuniform scale issue in clustered chips for better fine detection.
		
		\item Achieves state-of-the-art performance on three representative aerial image datasets including VisDrone~\cite{zhu2018vision}, UAVDT~\cite{du2018unmanned}, DOTA~\cite{xia2018dota} with less computation.
		
	\end{enumerate}
	
	The rest of this paper is organized as follows. Section~\ref{related_work} briefly reviews the related works. In Section~\ref{method}, we describe the proposed approach in details. Experimental results are shown in Section~\ref{exp}, followed by the conclusion in Section~\ref{sec:con}.


	\section{Related work}
	\label{related_work}
	
	Object detection has been extensively explored in recent decades with a huge amount of literature. In the following, we first review three lines of works that are the most relevant to ours, and then highlight the differences of CLusDet with existing approaches.
	
	\noindent
	{\bf Generic Object Detection.} Inspired by the success in image recognition~\cite{krizhevsky2012imagenet}, deep convolutional neural networks (CNNs) have been dominated in object detection. According to the detection pipeline, existing detectors can roughly be categorized into two types: region-based detectors and region-free detectors. The region-based detectors separate detection into two steps including proposal extraction and object detection. In the first stage, the search space for detection is significantly reduced through extracting candidate regions (\ie, proposals). In the second stage, these proposals are further classified into specific categories. Representatives of region-based detectors include R-CNN~\cite{girshick2014rich}, Fast/er R-CNN~\cite{girshick2015fast,ren2015faster}, Mask R-CNN~\cite{he2017mask} and Cascade R-CNN~\cite{cai2018cascade}. On the contrary, the region-free detectors, such as SSD~\cite{liu2016ssd} YOLO~\cite{redmon2016you}, YOLO9000~\cite{redmon2017yolo9000}, RetinaNet~\cite{lin2017focal} and RefineDet~\cite{zhang2018single}, perform detection without region proposal, which leads to high efficiency at the sacrifice of accuracy.
	
	Despite excellent performance on natural images (\eg, 500$\times$400 images in PASCAL VOC~\cite{Everingham15} and 600$\times$400 images in MS COCO~\cite{lin2014microsoft}), these generic detectors are degenerated when applied on high-resolution aerial images (\eg, 2,000$\times$1,500 images in VisDrone~\cite{zhu2018vision}, and even larger in UAV captured imagery~\cite{LiangTLBCB12fusion}). Note that detection in high resolution imagery recently has gained an increasing amount of research attention~\cite{Sun&etal19arXivHRR}.
	
	\noindent
	{\bf Aerial Image Detection.} Compared to detection in natural images, detection in aerial image is more challenging because (1) objects have small scales relative to the high-resolution aerial images and (2) targets are sparse and nonuniform and concentrated in certain regions. Since this work is focused on deep learning, we only review some relevant works using deep neural networks for aerial image detection. In~\cite{vsevo2016convolutional}, a simple CNNs based approach is presented for automatic detection in aerial images. The method in~\cite{audebert2017segment} integrates detection in aerial images with semantic segmentation to improve performance. In~\cite{sommer2017fast}, the authors directly extend the Fast/er R-CNN ~\cite{girshick2015fast,ren2015faster} for vehicle detection in aerial images. The work of~\cite{deng2017toward} proposes a coupled region-based CNNs for aerial vehicle detection. The approach of~\cite{ding2018learning} investigates the problem of misalignment between Region of Interests (RoI) and objects in aerial image detection, and introduces a ROI transformer to address this issue. The algorithm in~\cite{zhang2019scale} presents a scale adaptive proposal network for object detection in aerial images.
	
	\noindent
	{\bf Region Search in Detection.} The strategy of region search is commonly adopted in detection to handle small objects. The approach of~\cite{lu2016adaptive} proposes to adaptively direct computational resources to sub-regions where objects are sparse and small. The work of~\cite{NIPS2012_4717} introduces a context driven search method to efficiently localize the regions containing a specific class of object. In~\cite{chen2016object}, the authors propose to dynamically explore the search space in proposal-based object detection by learning contextual relations. The method in~\cite{gao2018dynamic} proposes to leverage reinforcement learning to sequentially select regions for detection at higher resolution scale. In a more specific domain, vehicle detection in wide aerial motion imagery (WAMI), the work of~\cite{lalonde2018clusternet} suggests a two-stage spatial-temporal convolutional neural networks to detect vehicles from a sequence of WAMI.
	
	{\bf Our Approach.} In this paper, we aim at solving two aforementioned challenges for aerial image detection. Our approach is related to but different from the previous region search based detectors (\eg,~\cite{lu2016adaptive,gao2018dynamic}), which partitions high-resolution images into small uniforms chips for detection. In contrast, our solution first predicts cluster regions in the images, and then extract these clustered regions for fine detection, leading to significant reduction of the computation cost. Although the method in~\cite{lalonde2018clusternet} also performs detection on chips that potentially contain objects, our approach significantly differs from it. In~\cite{lalonde2018clusternet}, the obtained chips are directly resized to fit the detector for subsequent detection. On the contrary, inspired by the observation in~\cite{singh2018analysis} that objects with extreme scales may deteriorate the detection performance, we propose a ScaleNet to alleviate this issue, resulting in improvement in fine detection on each chip.

	\section{Clustered Detection (ClusDet) Network}
	\label{method}
	
	\subsection{Overview}
	As shown in Fig.~\ref{fig:cluster_aware_scheme}, detection of an aerial image consists of three stages: cluster region extraction, fine detection on cluster chips and fusion of detection results. In specific, after the feature extraction of an aerial image, CPNet takes as input the feature maps and outputs the clustered regions. In order to avoid processing too many cluster chips, we propose an iterative cluster merging (ICM) module to reduce the noisy cluster chips. Afterwards, the cluster chips as well as the initial detection results on global image are fed into the ScaleNet to estimate an appropriate scale for the objects in cluster chips. With the scale information, the cluster chips are rescaled for fine detection with DetecNet. The final detection is obtained by fusing the detection results of each cluster chip and global image with standard non-maximum suppression (NMS).
	
	\subsection{Cluster Region Extraction}
	
	Cluster region extraction consists of two steps: initial cluster generation using cluster proposal sub-network (CPNet) and cluster reduction with iterative cluster merging (ICM).
	
	\subsubsection{Cluster Proposal Sub-network (CPNet)}
	
	The core of the cluster region extraction is the cluster proposal sub-network (CPNet). CPNet works on the high-level feature maps of an aerial image, and aims at predicting the locations and scales of clusters\footnote{In this work, a cluster in aerial images is defined by a rectangle region containing at least three objects.}. Motivated by the region  proposal networks (RPN)~\cite{ren2015faster}, we formulate CPNet as a block of fully convolutional networks. In specific, CPNet takes as input the high-level feature maps from feature extraction backbone, and utilizes two subnets for regression and classification, respectively. Although our CPNet shares the similar idea with RPN, they are different. RPN is used to propose the candidate regions of objects, while CPNet aims at proposing the candidate regions of clusters. Compared to the object proposal, the size of cluster is much larger, and thus CPNet needs a larger receptive field than that of RPN. For this reason, we attach CPNet on the top of the feature extraction backbone.
	\begin{figure}[t]
		\centering
		\begin{subfigure}[b]{0.5\linewidth}
			
			\includegraphics[width=\linewidth]{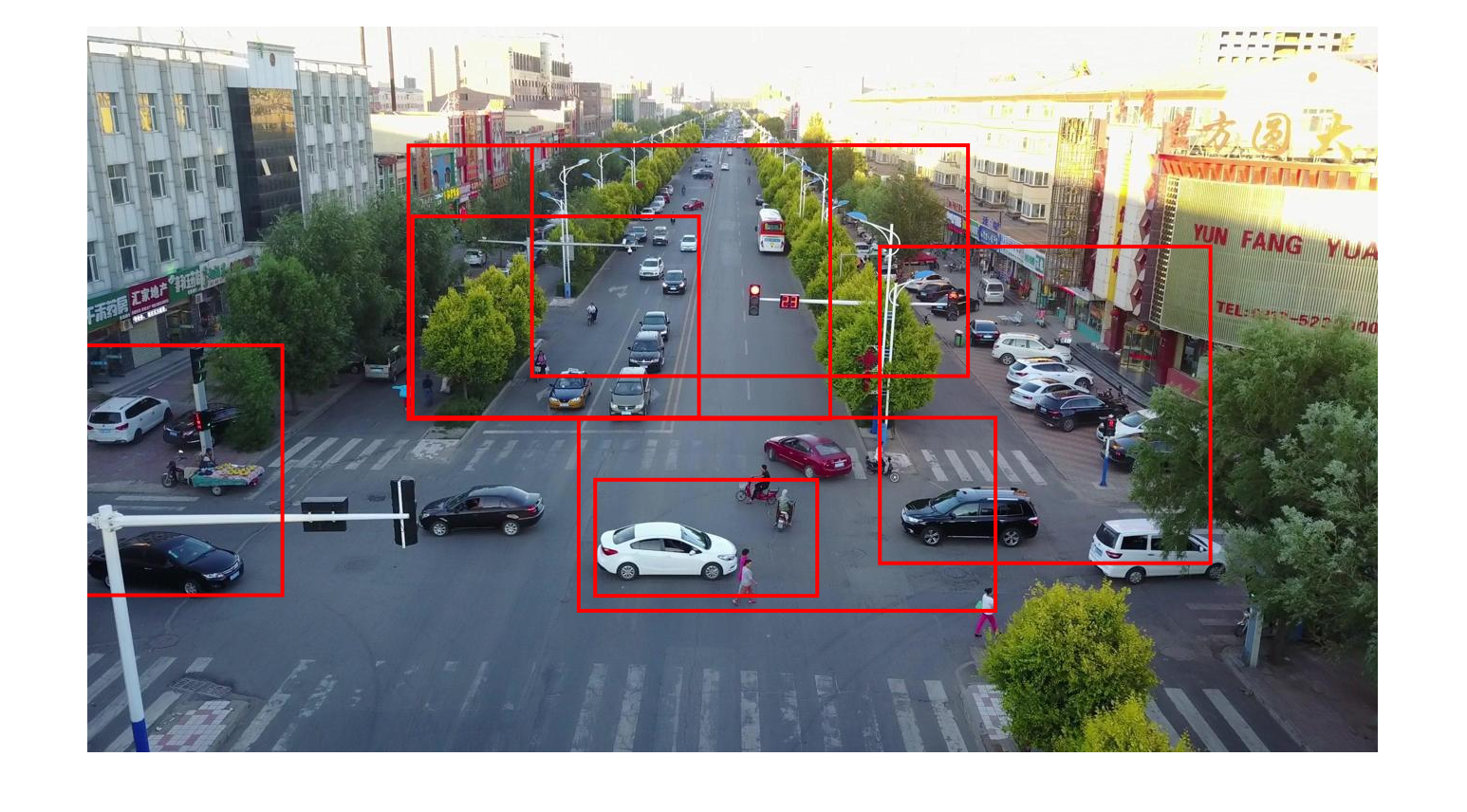}
			\caption{cluster detections}
			\label{fig:clusterdets}
		\end{subfigure}\hspace*{-2mm}
		~
		\begin{subfigure}[b]{0.5\linewidth}
			\includegraphics[width=\linewidth]{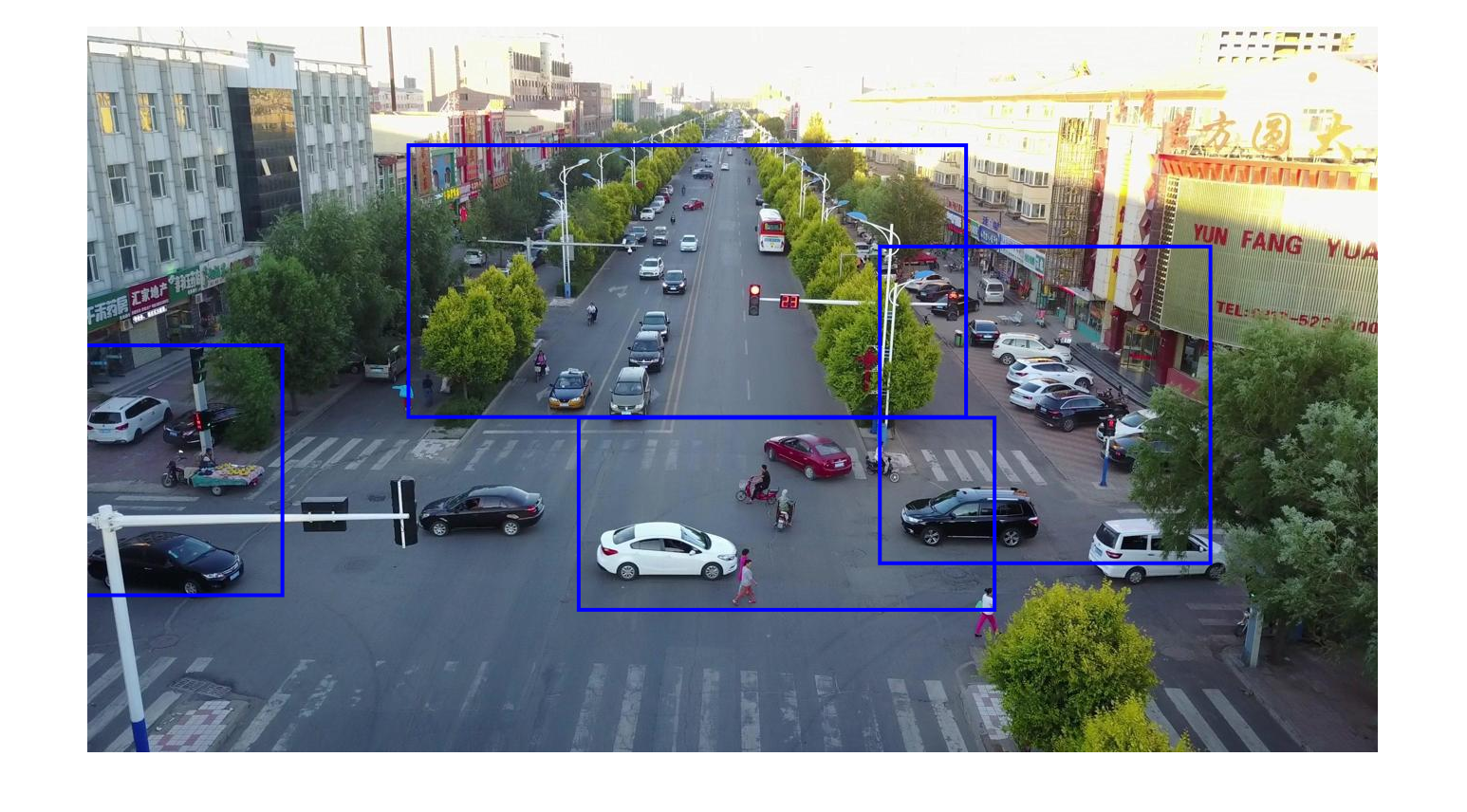}
			\caption{cluster detections + ICM}
			\label{fig:clustersdets-M}
		\end{subfigure}\hspace*{-2mm}
		\caption{Illustration of merging of cluster detections. The red boxes are the cluster detections from CPNet. The blue boxes represent clusters after iterative cluster merge (ICM).}
		\label{icm}
	\end{figure}

	It is worth noting that the learning of CPNet is a supervised process. However, none of existing public datasets provide groundtruth for clusters. In this work, we adopt a simple strategy to generate the required groundtruth of clusters for training CPNet. We refer the readers to supplementary material for details in generating cluster groundtruth.
	
	\subsubsection{Iterative Cluster Merging (ICM)}
	
	As shown in Fig.~\ref{icm} (a), we observe that the initial clusters produced by CPNet are dense and messy. These dense and messy cluster regions are difficult to be directly leveraged for fine detection because of their high overlap and large size,  resulting in extremely heavy computation burden in practice. To solve this problem, we present a simple yet effective iterative cluster merging (ICM) module to clean up clusters.
	Let $\mathcal{B}=\{B_i\}_{i=1}^{N_{\mathcal{B}}}$ represent the set of $N_{\mathcal{B}}$ cluster bounding boxes detected by CPNet, and $\mathcal{R}=\{R_i\}_{i=1}^{N_{\mathcal{B}}}$ denote the corresponding cluster classification scores. With a pre-defined overlap threshold $\tau_{\mathrm{op}}$ and maximum number $N_{\max}$ of clusters after merging, we can obtain the merged cluster set $\mathcal{B'}=\{B'_i\}_{i=1}^{N_{\mathcal{B'}}}$ with $N'_{\mathcal{B}}$ clusters with Alg.~\ref{IMCB}. 
	
	Briefly speaking, we first find the $B_i$ with highest score, then select the clusters whose overlaps with $B_i$ are larger than the threshold $\tau_{\mathrm{op}}$ to merge with $B_i$. All the merged clusters are removed. Afterwards, we repeat the aforementioned process until $\mathcal{B}$ is empty. All the processes mentioned above correspond to the non-max merging (NMM) in Alg.~\ref{IMCB}. We conduct the NMM several times until the preset $N_{\max}$ is reached. For the details of the NMM, the readers are referred to supplementary material.      
	Fig.~\ref{icm} (b) demonstrates the final merged clusters, showing that the proposed ICM module is able to effectively merge the dense and messy clusters.

	\begin{algorithm}[t]
		\SetAlgoLined
		\KwIn{Initial cluster bounding boxes $\mathcal{B}=\{B_i\}_{i=1}^{N_{\mathcal{B}}}$,\\ initial cluster scores $\mathcal{R}=\{R_i\}_{i=1}^{N_{\mathcal{B}}}$, threshold $\tau_{\mathrm{op}}$ and maximum number of merged clusters $N_{\max}$;}
		\KwOut{Merged clusters $\mathcal{B'}=\{B'_i\}_{i=1}^{N_{\mathcal{B'}}}$;}
		\Begin{
			$\mathcal{B'} \leftarrow \mathcal{B}$\;
			\While{$|\mathcal{B'}|$ $>$ $N_{\max}$}
			{
				$\mathcal{B'}$, $\mathcal{R'}$ $\leftarrow$ NMM$(\mathcal{B},\mathcal{R},\tau_{\mathrm{op}}$)\\ 
				\eIf{$|N'_{\mathcal{B}}|$ $=$ $|N_{\mathcal{B}}|$}
				{break\;}
				{$\mathcal{B}\leftarrow \mathcal{B'}$; $\mathcal{R} \leftarrow \mathcal{R'}$	\;
				}	
			}
			$\mathcal{B''}\leftarrow \{\}$\;
			\For{$i \leq \min(N_{\max},|\mathcal{B'}|)$}
			{
				$\mathcal{B''} \leftarrow \mathcal{B''} \cup \{B'_{i}\}$\;
			}
			$\mathcal{B'}\leftarrow \mathcal{B''}$\;	
		}
		\caption{Iterative Cluster Merging (ICM)}
		\label{IMCB}
		
	\end{algorithm}
	
	
	

	\subsection{Fine Detection on Cluster Chip}
	
	After obtaining the cluster chips, a dedicated detector is utilized to perform fine detection on these chips. Unlike in existing approaches~\cite{lu2016adaptive,lalonde2018clusternet,gao2018dynamic} that directly resize these chips for detection, we present a scale estimation sub-network (ScaleNet) to estimate the scales of objects in chips, which avoids extreme scales of objects degrading detection performance. Based on the estimated scales, ClusDet performs partition and padding (PP) operations on each chip for detection.
	
	\subsubsection{Scale Estimation Sub-network (ScaleNet)}
	
\begin{figure}[t]
		\centering
		\includegraphics[width=\linewidth]{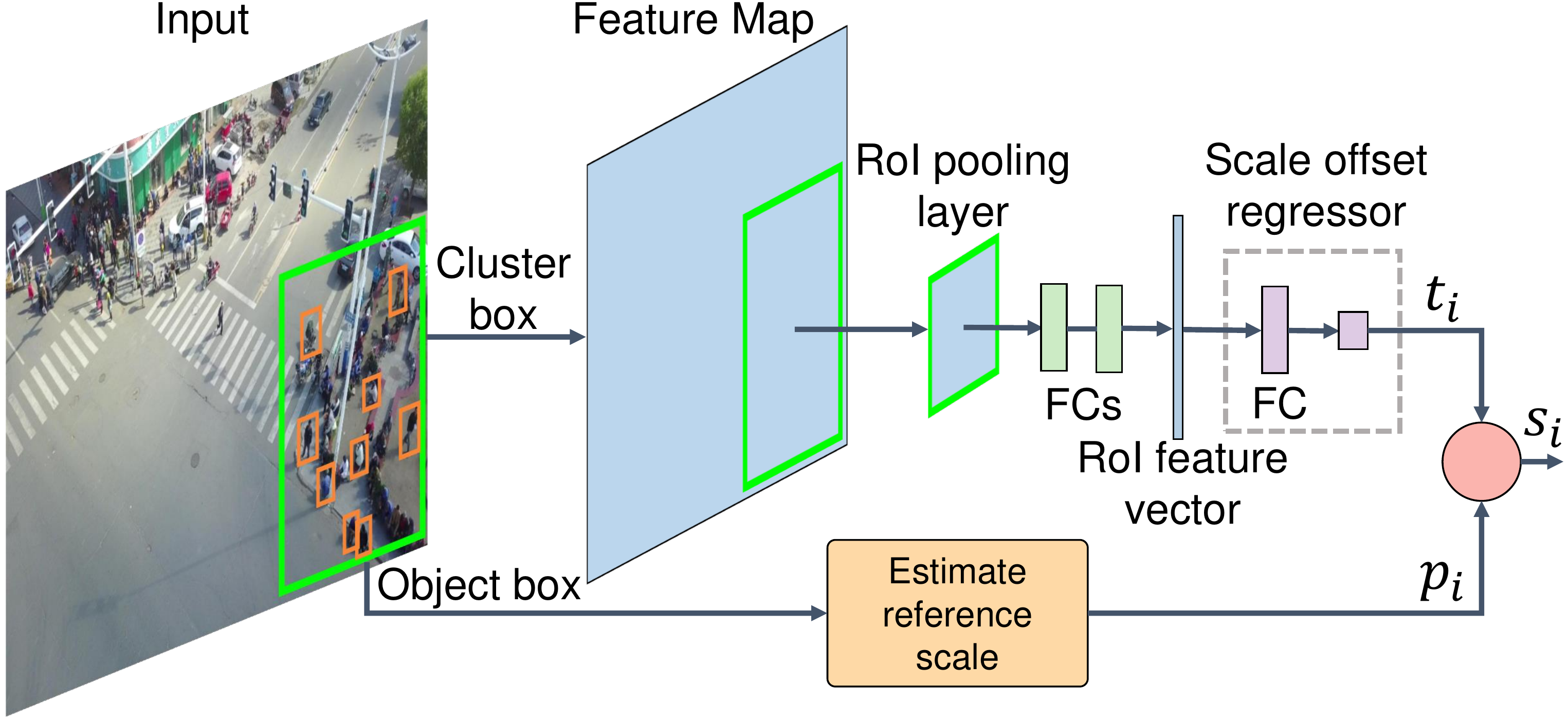}
		\caption{The architecture of the scale estimation network (ScaleNet). The cluster detections are projected to feature map space. Each cluster is pooled into a fixed-size feature map and mapped into a feature vector by fully connected layers (FCs). The network has an output per cluster, \ie, the scale regression offset.}
		\label{scalenet}
	\end{figure}

	We regard scale estimation as a regression problem and formulate ScaleNet using a bunch of fully connected networks. As shown in Fig.~\ref{scalenet}, ScaleNet receives three inputs including the feature maps extracted from network backbone, cluster bounding boxes and initial detection results on global image, and outputs a relative scale offset for objects in the cluster chip. Here, the initial detection results are obtained from the detection subnet.
	
	Let $t^*_i=(p_i-s^*_i)/p_i$ be the relative scale offset for cluster $i$, where $p_i$ and $s^*_i$ represent the reference scale of the detected objects and the average scale of the groundtruth boxes in cluster $i$, respectively. Thus, the loss of the ScaleNet can be mathematically defined as
	\begin{equation} \label{eu_eqn}
	\mathcal{L}(\{t_i\})=\frac{1}{M}\sum_{i}^{M} \ell_\mathrm{reg}(t_i,t_i^*)
	\end{equation}
	where $t_i=(p_i-s_i)/p_i$ is the estimated relative scale offset, $s_i$ is the estimated scale, and $M$ is the number of cluster boxes. The $\ell_\mathrm{reg}$ is a smoothly $\ell_1$ loss function~\cite{girshick2015fast}.
	
	\subsubsection{Partition and Padding (PP)}
	
\begin{figure}[t]
		\centering
		\includegraphics[width=\linewidth]{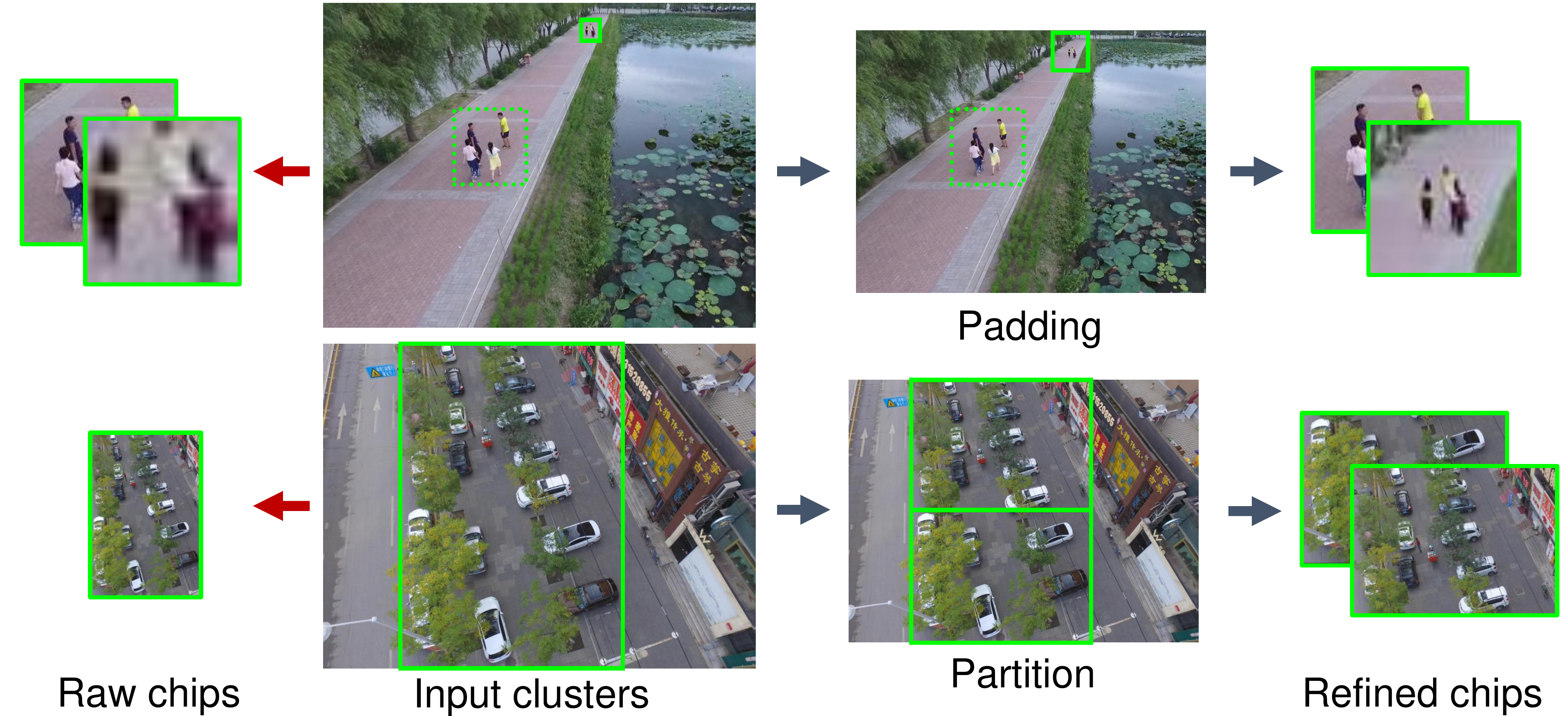}
		\caption{Illustration of the partition and padding (PP) process. The raw chips and refined chips are the input of detector without and with using PP, respectively.}
		\label{pp}
	\end{figure}

	The partition and padding (PP) operations are utilized to ensure that the scales of objects are within a reasonable range. Given the cluster bounding box $B_i$, the corresponding estimated object scale $S_i$ and the input size $S_{in}$ of a detector, we can estimate the object scale in the input space of the detector $S_i^{in}=S_i \times \frac{S_{in}}{S_i} $. If the scale $S_i^{in}$ is larger than a certain range, the cluster is padded proportionally, otherwise it is partitioned into two equal chips. Note that detections in the padded region are ignored in final detection. The visualization of the process is in Fig.~\ref{pp}. 
	The specific scale range setting is discussed in Section~\ref{exp}. 
	
	After rescaling the cluster chip, a dedicated baseline detection network (DetecNet) performs fine object detection. The architecture of the DetecNet can be any state-of-the-art detectors. The backbone of the detector can be any standard backbone networks, e.g.,  VGG~\cite{simonyan2014very}, ResNet~\cite{he2016deep}, ResNeXt~\cite{xie2017aggregated}.
	
	\subsection{Final Detection with Local-Global Fusion}
	
\begin{figure}[t]
		\centering
		\includegraphics[width=\linewidth]{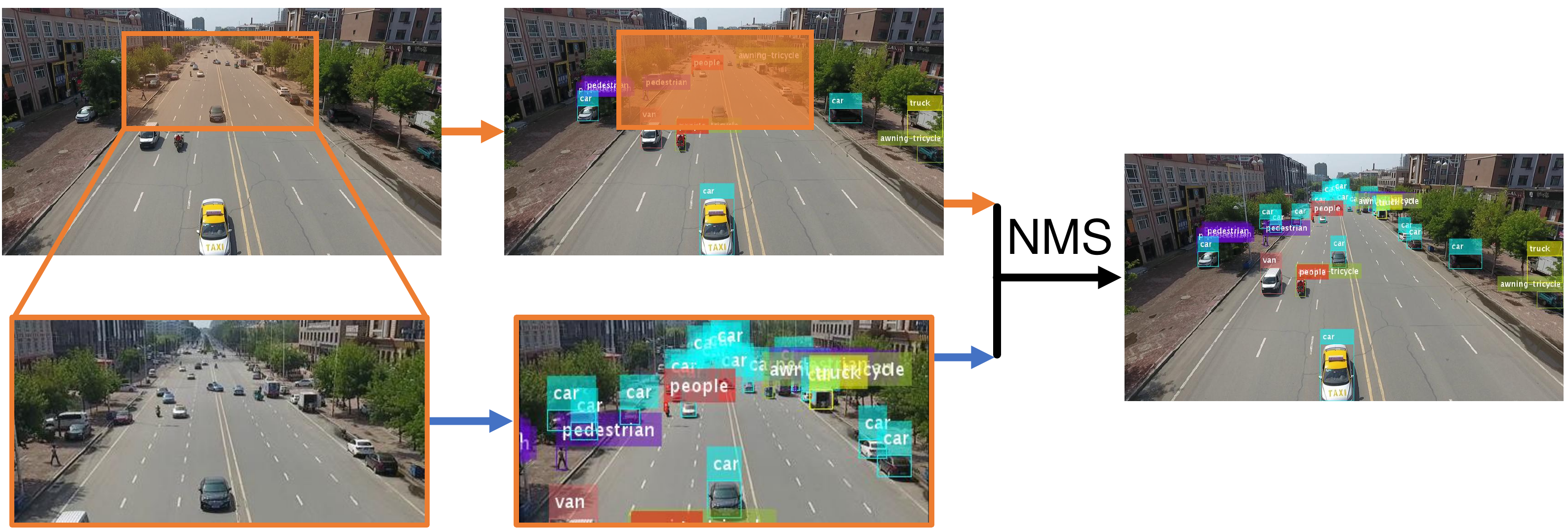}
		\caption{The illustration of fusing detections from whole images and cluster chips.The object detections in orange region from whole image are eliminated when applying fusion operation.}
		\label{fig:fusion}
	\end{figure}

	The final detection of an aerial image is obtained by fusing the local detection results of cluster chips and global detection results of the whole image with the standard NMS post-processing (see Fig.~\ref{fig:fusion}). The local detection results are obtained through the proposed approach mentioned above, and the global detection results are derived from detection subnet (Fig.~\ref{fig:cluster_aware_scheme}).  It is worth noting that any existing modern detectors can be used for global detection.
	
	\begin{table*}[t]
		\centering
		\footnotesize
		\caption{The ablation study on VisDrone dataset. The `c' denotes EIP cropped images. The 'ca' indicates cluster-aware cropped images. The `o' indicates the original validation data. The $\#$img is the number of images forwarded to detector. The `s', `m', and `l' represent small, medium, and large, respectively. The inference time is measured on a GTX 1080 Ti.}
		\begin{tabular}{c|c|c|c|c|c|c|c|c|c|c}
			\hline\hline
			Methods&backbone&test data&$\#$img& $AP$&$AP_{50}$&$AP_{75}$&$AP_{s}$&$AP_{m}$&$AP_{l}$&s/img (GPU)\\
			\hline\hline
			FRCNN\cite{ren2015faster}+FPN\cite{lin2017feature}&ResNet50&o&548&21.4&40.7&19.9&11.7&33.9&54.7&0.055\\ 
			\hline
			FRCNN\cite{ren2015faster}+FPN\cite{lin2017feature}&ResNet101&o&548&21.4&40.7&20.3&11.6&33.9&54.9&0.074\\ 
			\hline
			FRCNN\cite{ren2015faster}+FPN\cite{lin2017feature}&ResNeXt101&o&548&21.8&41.8&20.1&11.9&34.8&55.5&0.156\\
			\hline
			FRCNN\cite{ren2015faster}+FPN\cite{lin2017feature}+EIP&ResNet50&c&3,288 &21.1&44.0&18.1&14.4&30.9&30.0&0.330\\ 
			\hline
			FRCNN\cite{ren2015faster}+FPN\cite{lin2017feature}+EIP&ResNet101&c&3,288 &23.5&46.1&21.1&17.1&33.9&29.1&0.444\\
			\hline
			FRCNN\cite{ren2015faster}+FPN\cite{lin2017feature}+EIP&ResNeXt101&c&3,288 &24.4&47.8&21.8&17.8&34.8&34.3&0.936\\
			
			\hline
			\hline
			DetecNet+CPNet &ResNet50&o+ca&1,945 &25.6&47.9&24.3&16.2&38.4&53.7&0.195\\
			\hline
			DetecNet+CPNet &ResNet101&o+ca&1,945 &25.3&47.4&23.8&15.6&38.1&\textbf{54.6}&0.262\\
			\hline
			DetecNet+CPNet &ResNeXt101&o+ca&1,945&27.6&51.2&26.2&17.5&\textbf{41.0}&54.2&0.554\\
			\hline
			DetecNet+CPNet+ScaleNet &ResNet50&o+ca&2,716&26.7&50.6&24.7&17.6&38.9&51.4&0.273\\
			\hline
			DetecNet+CPNet+ScaleNet &ResNet101&o+ca&2,716&26.7&50.4&25.2&17.2&39.3&54.9&0.366\\
			\hline
			DetecNet+CPNet+ScaleNet &ResNeXt101&o+ca&2,716&\textbf{28.4}&\textbf{53.2}&\textbf{26.4}&\textbf{19.1}&40.8&54.4&0.773\\
			\hline
			
		\end{tabular}
		\label{table:albation}
	\end{table*}
	
	\section{Experiments}
	\label{exp}
	\subsection{Implementation Details}
	We implement ClusDet based on the publicly available Detectron~\cite{Detectron2018} and Caffe2. The Faster R-CNN (FRCNN)~\cite{ren2015faster} with Feature Pyramid Network (FPN)~\cite{lin2017feature} are adopted as the baseline detection network (DetecNet). 
	The architecture of the CPNet is implemented with a $5 \times 5 $ convolutional layer followed by two sibling $1 \times 1$ convolutional layers (for regression and classification, respectively).
	In ScaleNet, the FC layers to convert feature map into feature vector are with size of 1024; The size of FC layers in the scale offset regressor are 1024 and 1 respectively.
	The IoU threshold for merging clusters in NMM process is set to 0.7.
	Following the definition in the COCO\cite{lin2014microsoft} dataset, the object scale range in cluster chip partition and padding is set to $[70,280]$ pixels. 
	
	\textbf{Training phase.} The input size of the detector is set to $600 \times 1,000$ pixels on the VisDrone~\cite{zhu2018vision} and UAVDT~\cite{du2018unmanned} datasets and $1,000 \times 1,000$ pixels on the DOTA~\cite{xia2018dota} dataset. On the three datasets, the training data  is augmented by dividing images into chips. On the VisDrone~\cite{zhu2018vision} and UAVDT~\cite{du2018unmanned} datasets, each image is uniformly divided into 6 and 4 chips without overlap. 
	The reason of setting a specific number of chips is that the size of cropped chip can be similar with that in COCO~\cite{lin2014microsoft} dataset.  
	On the DOTA~\cite{xia2018dota} dataset, we use the tool provided by the authors to divide the images.
	When training model on the VisDrone~\cite{zhu2018vision} and
	UAVDT~\cite{du2018unmanned} datasets by using 2 GPUs, we set the base learning rate to 0.005 and total iteration to 140k. After the first 120k iterations, the learning rate decreases to 0.0005. Then, we train the model for 100k iterations before lowering the learning rate to 0.00005. A momentum of 0.9 and parameter decay of 0.0005 (on weights and biases) are used. On the DOTA~\cite{xia2018dota} dataset, the base learning and the total iterations are set to 0.005 and 40k, respectively. The learning rate is deceased by a factor of 0.1 after 30k and 35k iterations.
	
	\textbf{Test phase.} The input size of detector is the same with that in training phase whenever not specified.  
	The maximum number of clusters (TopN) in cluster chip generation is empirically set to 3 on VisDrone~\cite{zhu2018vision}, 2 on UAVDT~\cite{du2018unmanned}, and 5 on the DOTA~\cite{xia2018dota}. In fusing detection, the threshold of the standard non-max suppression (NMS) is set to 0.5 in all datasets. The final detection number is set to 500.   
	
	\subsection{Datasets}
	To validate the effectiveness of the proposed method, we conduct extensive experiments on three publicly accessible datasets: VisDrone~\cite{zhu2018vision}, UAVDT~\cite{du2018unmanned}, and DOTA~\cite{xia2018dota}.

	\textbf{VisDrone.} The dataset consists of 10, 209 images (6,471 for training, 548 for validation, 3,190 for testing) with rich annotations on ten categories of objects. The image scale of the dataset is about $2,000\times1,500$ pixels. Since the evaluation server is closed now, we cannot test our method on the test dataset.
	Therefore, the validation dataset is used as test dataset to evaluate our method.
	
	
	
	\textbf{UAVDT.} The UAVDT~\cite{du2018unmanned}] dataset contains 23,258 images of training data and 15,069 images of test data. The resolution of the image is about $1,080\times540$ pixels. The dataset is acquired with an UAV platform at a number of locations 
	in urban areas. The categories of the annotated objects are car, bus, and truck.
	\begin{figure}[t]
		\centering
		\includegraphics[width=\linewidth]{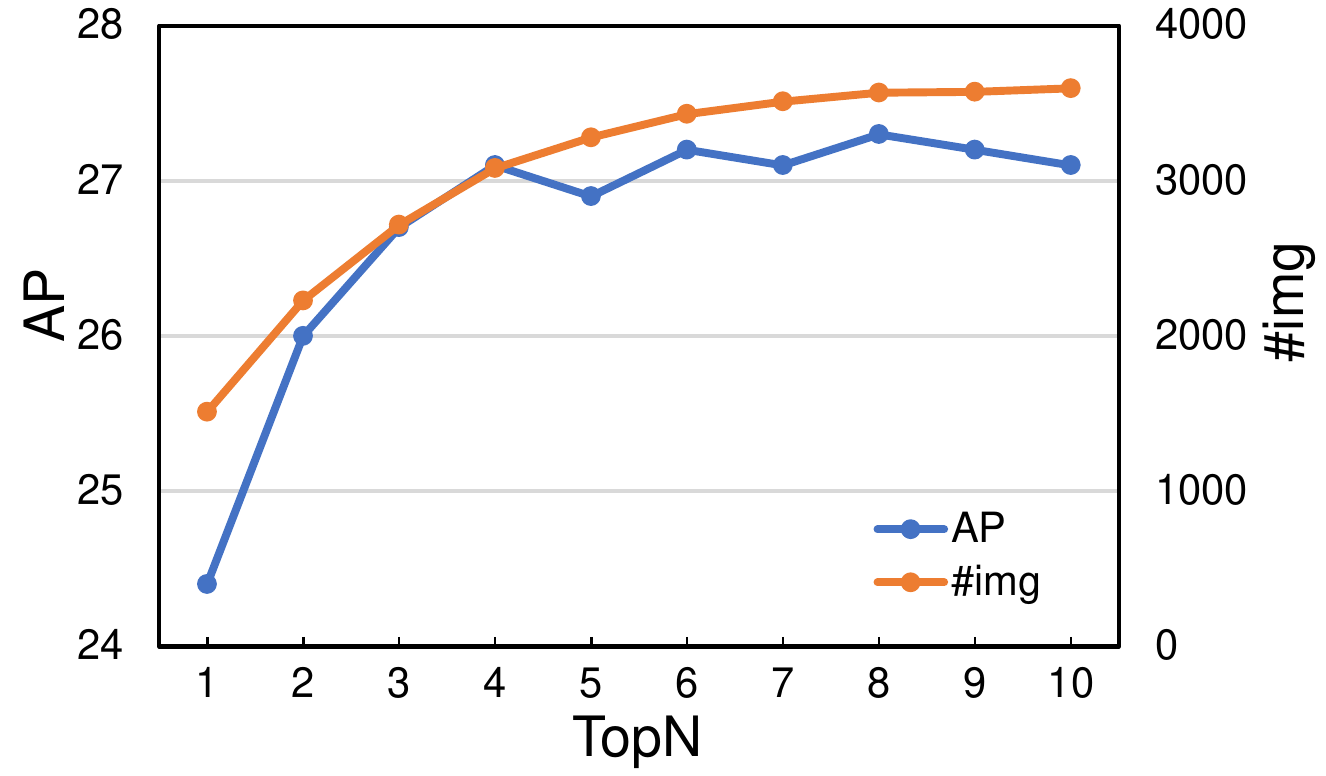}
		\caption{The AP and number of forwarded images over different settings of TopN in ClusDet.}
		\label{fig:topN}
	\end{figure}
	
	\textbf{DOTA.} The dataset is collected from multiple
	sensors and platforms (e.g. Google Earth) with multiple resolutions (800$\times$800 through 4,000$\times$4,000 pixels) at multiple cities. Fifteen categories are chosen and annotated.
	Considering that ClusDet is based on the cluster characteristic of the objects in aerial image, some categories in the dataset are not suitable for ClusDet, e.g., roundabout, bridge. Thus, we only choose the images with movable objects in the dataset to evaluate our method, i.e., plane, ship, large vehicle, small vehicle, and helicopter, Thus, the training and validation data contain 920 images and 285 images, respectively.  
	
	\subsection{Compared Methods}
	We compare our ClusDet with evenly image partition (EIP) method on all datasets. 
	On some datasets if the EIP is not provided, we implement it according to the property of the datasets. In addition, we also compare our method with representative state-of-the-art methods on all datasets.
	
	\begin{table}[t]
		\centering
		\footnotesize
		\caption{The detection performance on VisDrone validation dataset. The $\boldsymbol{\star}$ denotes the multi-scale inference and bounding box voting are utilized in test phase.}
		\begin{tabular}{@{\hskip 1mm}c@{\hskip 1mm}|@{\hskip 1mm}c@{\hskip 1mm}|c|c|c}
			\hline\hline
			Methods&backbone& $AP$&$AP_{50}$&$AP_{75}$\\
			\hline\hline
			\makecell{RetinaNet\cite{lin2017focal}+FPN\cite{lin2017feature}}&ResNet50&13.9&23.0&14.9\\
			\hline
			\makecell{RetinaNet\cite{lin2017focal}+FPN\cite{lin2017feature}}&ResNet101&14.1&23.4&14.9\\
			\hline
			\makecell{RetinaNet\cite{lin2017focal}+FPN\cite{lin2017feature}}&ResNeXt101&14.4&24.1&15.5\\
			\hline
			\makecell{FRCNN\cite{ren2015faster}+FPN\cite{lin2017feature}} &ResNet50&21.4&40.7&19.9\\
			\hline
			
			\makecell{FRCNN\cite{ren2015faster}+FPN\cite{lin2017feature}} &ResNet101&21.4&40.7&20.3\\
			\hline
			\makecell{FRCNN\cite{ren2015faster}+FPN\cite{lin2017feature}}&ResNeXt101&21.8&41.8&20.1\\
			\hline
			\makecell{FRCNN\cite{ren2015faster}+FPN\cite{lin2017feature} $\boldsymbol{\star}$}&ResNeXt101&28.7&51.8&27.7\\
			\hline
			\makecell{FRCNN\cite{ren2015faster}+FPN\cite{lin2017feature}+EIP}&ResNet50&21.1&44.0&18.1\\
			\hline
			\makecell{FRCNN\cite{ren2015faster}+FPN\cite{lin2017feature}+EIP}&ResNet101&23.5&46.1&21.1\\
			\hline
			
			\makecell{FRCNN\cite{ren2015faster}+FPN\cite{lin2017feature}+EIP}&ResNeXt101&24.4&47.8&21.8\\
			\hline
			
			\makecell{FRCNN\cite{ren2015faster}+FPN\cite{lin2017feature}+EIP$ \boldsymbol{\star}$}&ResNeXt101&25.7&48.4&24.1\\
			\hline
			\hline
			\makecell{ClusDet} &ResNet50&26.7&50.6&24.7\\
			\hline
			\makecell{ClusDet} &ResNet101&26.7&50.4&25.2\\
			\hline
			
			\makecell{ClusDet} &ResNeXt101&28.4&53.2&26.4\\
			\hline
			\makecell{ClusDet $\boldsymbol{\star}$} &ResNeXt101&\textbf{32.4}&\textbf{56.2}&\textbf{31.6}\\
			\hline
		\end{tabular}
		\label{table:VisDrone_AP}
	\end{table}
	\begin{table*}[t]
		\centering
		\footnotesize
		\caption{The detection performance of the baselines and proposed method on the UAVDT~\cite{du2018unmanned} dataset.}
		\begin{tabular}{c|c|c|c|c|c|c|c|c}
			\hline\hline		
			Methods&backbone&$\#$img& $AP$&$AP_{50}$&$AP_{75}$&$AP_{s}$&$AP_{m}$&$AP_{l}$
			\\
			\hline\hline
			R-FCN\cite{dai2016r}&ResNet50&15,069&7.0&17.5&3.9&4.4&14.7
			&12.1
			\\ 
			\hline
			SSD\cite{liu2016ssd}&N/A&15,069&9.3&21.4&6.7&7.1&17.1&12.0
			\\ 
			\hline
			RON\cite{kong2017ron}&N/A&15,069&5.0&15.9&1.7&2.9&12.7&11.2
			\\ 
			\hline
			FRCNN\cite{ren2015faster}&VGG&15,069&5.8&17.4&2.5&3.8&12.3&9.4
			\\ 
			\hline
			FRCNN\cite{ren2015faster}+FPN\cite{lin2017feature}&ResNet50&15,069&11.0&23.4&8.4&8.1&20.2&26.5
			\\ 
			\hline
			FRCNN\cite{ren2015faster}+FPN\cite{lin2017feature}+EIP&ResNet50&60,276&6.6&16.8&3.4&5.2&13.0&17.2
			\\ 
			\hline
			\hline
			ClusDet &ResNet50&25,427&\textbf{13.7}&\textbf{26.5}&\textbf{12.5}&\textbf{9.1}&\textbf{25.1}&\textbf{31.2}
			\\
			\hline
		\end{tabular}
		\label{table:UAVDT_AP}
	\end{table*}
	\begin{table*}[t]
		\centering 
		\footnotesize
		\caption{The detection performance of the baselines and proposed method on DOTA~\cite{xia2018dota} dataset.}
		\begin{tabular}{c|c|c|c|c|c|c|c|c}
			\hline\hline
			Methods&backbone&$\#$img& $AP$&$AP_{50}$&$AP_{75}$&$AP_{s}$&$AP_{m}$&$AP_{l}$\\
			\hline\hline
			RetinaNet\cite{lin2017focal}+FPN\cite{lin2017feature}+EIP&ResNet50&2,838 &24.9&41.5&27.4&9.9&32.7&30.1\\ 
			\hline
			RetinaNet\cite{lin2017focal}+FPN\cite{lin2017feature}+EIP&ResNet101&2,838 &27.1&44.4&30.1&10.6&34.8&33.7\\ 
			\hline
			RetinaNet\cite{lin2017focal}+FPN\cite{lin2017feature}+EIP&ResNeXt101&2,838 &27.4&44.7&29.8&10.5&35.8&32.8\\ 
			\hline
			FRCNN\cite{ren2015faster}+FPN\cite{lin2017feature}+EIP&ResNet50&2,838 &31.0&50.7&32.9&16.2&37.9&37.2\\ 
			\hline
			FRCNN\cite{ren2015faster}+FPN\cite{lin2017feature}+EIP&ResNet101&2,838 &31.5&50.4&36.6&16.0&38.5&38.1\\
			\hline
			\hline
			ClusDet&ResNet50&1,055 &\textbf{32.2}&47.6&\textbf{39.2}&\textbf{16.6}&32.0&\textbf{50.0}\\ 
			\hline
			ClusDet&ResNet101&1,055 &31.6&47.8&38.2&15.9&31.7&49.3\\ 
			\hline
			ClusDet&ResNeXt101&1,055 &31.4&47.1&37.4&17.3&32.0&45.4\\ 
			\hline
			
		\end{tabular}
		\label{table:DOTA_AP}
	\end{table*}
	\subsection{Evaluation Metric}
	Following the evaluation protocol on the COCO~\cite{lin2014microsoft} dataset,
	we use $AP$, $AP_{50}$, and $AP_{75}$ as the metrics to measure the precision. Specifically, $AP$ is computed by averaging over all categories. $AP_{50}$ and $AP_{75}$ are computed at the single IoU threshold 0.5 and 0.75 over all categories. The efficiency is measured by the number of images needed to be processed by the detector and the average time to process a global image and its chips in inference stage. In specific, the number of images refer to the summation of global images and cropped chips. In the subsequent experiments, the number of images is denoted as $\#img$. 
	
	\subsection{Ablation Study}
	To validate the contributions of the cluster detection and scale estimation to detection improvement, we conduct extensive experiments on VisDrone~\cite{zhu2018vision}. 
	
	In the following experiments, the input size of detector in the test phase is set to $600 \times 1,000$ pixels. 
	To validate if the proposed method can gain consistent improvement in performance under different backbone networks, we conduct experiments with three backbone networks:  ResNet-50~\cite{he2016deep}, ResNet-101~\cite{he2016deep}, and ResNeXt-101~\cite{xie2017aggregated}.
	
	\textbf{Effect of EIP.} The experimental results are listed in Table~\ref{table:albation}. We note that FRCNN~\cite{ren2015faster} performs inferior compared to that in COCO~\cite{lin2014microsoft} (AP=36.7). This is because the relative scale of object to image in VisDrone~\cite{zhu2018vision} is much smaller than that in COCO~\cite{lin2014microsoft}.  
	By applying EIP to the image, the performance of detectors are increased significantly, especially on small objects ($AP_{s}$). However, the number of images needed to be processed increases 6 times (3,288 vs 548). In addition, we note that although the overall performance $AP$ is improved by applying EIP, the performance of large scale objects ($AP_{l}$) is decreased. This is because the EIP truncates the large objects into pieces, which results in many false positives.
	
	\textbf{Effect of Cluster Detection.}  
	From Table~\ref{table:albation}, we note that the DetecNet+CPNet processes much less amount of images (1,945 vs 3,288) but achieves better performance than FRCNN~\cite{ren2015faster} plus EIP. This demonstrates that the CPNet not only selects the clustered regions to save computation resource but also implicitly encodes the prior context information to improve the performance. 
	In addition, we note that compared to EIP, the CPNet dose not reduce the performance of large objects ($AP_{l}$), this can be attributed to the CPNet, which introduces the spatial distribution information of the object into the ClusDet network so as to avoid truncating the large object.
	
	\textbf{Effect of Scale Estimation.}
	After integrating ScaleNet into CPNet and DetecNet, we note that the number of processed image increases to 2,716, this is because the PP module partitions some cluster chips into pieces. This mitigates the small scale problem when performing detection, such that the performance ($AP$) is improved to 26.7 on ResNet50~\cite{he2016deep} backbone network. In addition, we see that the ScaleNet improves the detection performance on all types of backbone networks. Particularly, the metric $AP_{50}$ is boosted by 2-3 points. In addition, the $AP_{s}$ is increased by 1.6 points even on very strong backbone, ResNeXt101~\cite{he2016deep}. This demonstrate that the ScaleNet does alleviate the scale problem to certain extent. 
	
	\textbf{The Effect of Hyperparameter TopN.}
	To fairly investigate the effect of TopN, we only change the setting in test phase, which avoids the influence by the amount of training data. From Fig.~\ref{fig:topN}, we see that after $TopN=4$, the number of processed images gradually increases, yet the AP dose not change too much and just fluctuates around $AP=27$. This means that a lot of cluster regions are repetitively computed when TopN is set to a high value. This observation also indicates that the cluster merge operation is critical to decrease the computation cost.

	\subsection{Quantitative Results}
	\textbf{VisDrone}
	The detection performance of the proposed method  and representative detectors, i.e., Faster RCNN~\cite{ren2015faster} and RetinaNet~\cite{lin2017focal}, is shown in Table~\ref{table:VisDrone_AP}. We note that our method outperforms the state-of-the-art methods by a large margin over various backbone settings. Besides, we observe that when testing the model using multi-scale setting (denoted by $\boldsymbol{\star}$), the performance is significantly boosted, except for the methods using EIP. This is because in multi-scale test, the cropped chips are resized to extremely large scale such that detectors output many false positives on background or local regions of objects.

	\textbf{UAVDT} 
	The experimental results on the UAVDT~\cite{du2018unmanned} dataset are displayed in Table~\ref{table:UAVDT_AP}. The performance of the compared methods, except for FRCNN~\cite{ren2015faster}+FPN~\cite{lin2017feature}, is computed using the experimental results provided in~\cite{du2018unmanned}. From the Table~\ref{table:UAVDT_AP}, we observe that applying EIP on test data dose not improve the performance. On the contrary, it dramatically decreases the performance (11.0 vs 6.1). The reason of this phenomenon is that the objects, i.e. vehicles, in the UAVDT always appear in the center of the image, while the EIP operation divides the objects into pieces such that the detector cannot correctly estimate the objects scale. Compared to FRCNN~\cite{ren2015faster}+FPN~\cite{lin2017feature} (FFPN), our ClusDet is superior to the FFPN and FFPN+EIP. The performance improvement mainly benefits from the different image crop operation. In our method, the image is cropped based on the clusters information, which is less likely to truncate numerous objects. The performance of detectors on UAVDT~\cite{du2018unmanned} is much lower than that on VisDrone~\cite{zhu2018visdrone}, which is caused by the extremely unbalanced data. 
	
	\textbf{DOTA}
	On the DOTA\cite{xia2018dota} dataset, our ClusDet achieves similar performance with state-of-the-art methods but processes dramatically less image chips. This is because the CPNet significantly reduces the number of chips for fine detection. Although our method does not outperform the state-of-the-art methods in term of the overall performance at low IoU ($AP_{50}$), it obtains higher $AP_{75}$ value, which indicates that our method can more precisely estimate the object scale.   
	Besides, we observe that the performance does not change too much when more complex backbone networks are adopted. This can be attributed to the limited training images. 
	Without a large amount of data, the complex model cannot achieve its superiority. 
	\section{Conclusion}\label{sec:con}
	We present a Clustered object Detection (ClusDet) network to unify object clustering and detection in an end-to-end framework. We show that ClusDet can successfully predict the clustered regions in images to significantly reduce the number of chips for detection so as to improve the efficiency. Moreover, we propose a cluster-based object scale estimation network to effectively detect the small object. In addition, we  experimentally demonstrate that the proposed ClusDet network implicitly models the prior context information to improve the detection precision.
	By extensive experiments, we show that our method obtains state-of-the-art performance on three public datasets.
	
	\vspace{0.1em}
	\noindent
	{\small {\bf Acknowledgement.} We sincerely appreciate anonymous reviewers for their helpful comments in improving the draft. This work is supported in part by US NSF Grants 1814745, 1407156 and 1350521.}
	
	{\small
		\bibliographystyle{ieee_fullname}
		\bibliography{egbib}
	}
	
\end{document}